\documentclass[conference, a4paper]{APSIPA2021}
\usepackage[fleqn]{amsmath}
\usepackage[pdftex]{graphicx}
\usepackage{multirow}
\usepackage{newtxmath}
\usepackage{booktabs}
\usepackage{mathtools}
\usepackage{algorithm}
\usepackage{algpseudocode}
\usepackage[backend=biber, style=ieee, citestyle=numeric-comp]{biblatex}
\usepackage{geometry}
\usepackage{fancyhdr}

\fancypagestyle{firststyle}{
  \fancyhf{}
  \fancyhead[C]{2023 Asia Pacific Signal and Information Processing Association
    Annual Summit and Conference (APSIPA ASC)}
}

\makeatletter
\let\MYcaption\@makecaption
\makeatother
\usepackage[hang,small,bf]{caption}
\usepackage[subrefformat=parens]{subcaption}
\makeatletter
\let\@makecaption\MYcaption
\makeatother

\begin{document}
\title{Kernel Random Projection Depth \\for Outlier Detection}
\author{
  \authorblockN{Akira Tamamori}
  \authorblockA{
    Department of Information Science, Aichi Institute of Technology, Aichi, Japan\\
    E-mail: akira-tamamori@aitech.ac.jp}
}
\maketitle

\thispagestyle{firststyle}
\pagestyle{fancy}               
\pagenumbering{gobble}          

\begin{abstract} This paper proposes an extension of Random Projection Depth
  (RPD) to cope with multiple modalities and non-convexity on data clouds. In
  the framework of the proposed method, the RPD is computed in a reproducing
  kernel Hilbert space. With the help of kernel principal component analysis,
  we expect that the proposed method can cope with the above multiple
  modalities and non-convexity. The experimental results demonstrate that the
  proposed method outperforms RPD and is comparable to other existing detection
  models on benchmark datasets regarding Area Under the Curves (AUCs) of
  Receiver Operating Characteristic (ROC).
\end{abstract}

\section{Introduction}
\begin{figure*}[t]
  \centering
  \includegraphics[width=\linewidth]{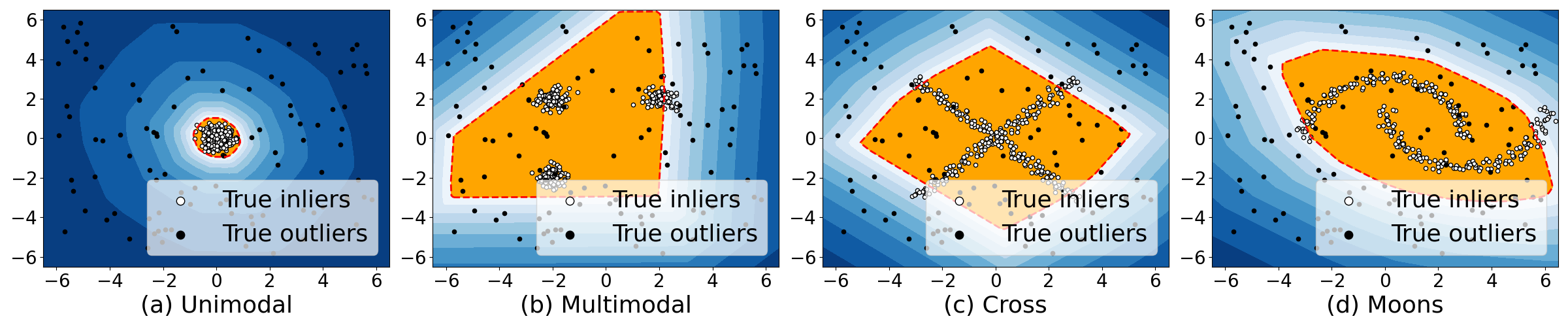}
  \caption{Contour plots of RPD.
    The decision boundary is indicated by the red dot line and
  the central region is colored orange.}
  \label{fig:rpo_datasets}
\end{figure*}

Outlier detection refers to ``the problem of finding patterns in data that do
not conform to expected normal behavior''~\cite{Chandola2009}. Outlier
detection has broad applicability and many application examples, such as fraud
detection, intrusion detection, disease detection, medical diagnosis, etc. In
supervised learning of classification problems, the authors have reported that
utilizing outlier detection for data cleaning, i.e., removing anomalous samples
as a preprocessing step, significantly improves the classification
accuracy~\cite{Smith2011}. Accordingly, outlier detection is one of the
important research topics in the field of data mining and machine learning.

Outlier detection algorithms often quantify the
\textit{outlyingness}~(outlierness) of points on a data cloud in a
multidimensional space. The opposite of outlyingness is the
depth~\cite{Zuo2000}. It measures the centrality of a data point, which defines
how close the point is to the center of the data cloud. In other words, the
depth introduces a center-outward ordering to the data cloud. Since we can
construct analogs of quantiles and ranks in multidimensional datasets from the
depth, many authors successfully have applied the depth function in related
research fields, such as robust estimation~\cite{Chen2004},
nonparametric tests~\cite{Chenouri2012},
cluster analysis~\cite{Hoberg2000},
discriminant analysis~\cite{Ghosh2005},
multivariate density estimation~\cite{Fraiman1997},
functional data analysis~\cite{Sara2009}, etc.
Authors proposed various depth functions in the literature
listed as follows: halfspace depth~\cite{Tukey1975},
simplicial volume depth~\cite{Oja1983}, zonoid
depth~\cite{Koshevoy1997},
spatial depth~\cite{Vardi2000}, projection
depth~\cite{Liu1992, Zuo2000}, Mahalanobis depth~\cite{mahalanobis1936,
  Zuo2000},
and so on.
Among these depth functions, it is known that the halfspace depth and projection
depth are favorable since they satisfy the desirable properties of the
statistical depth~\cite{Zuo2000}; affine invariance, maximality at the center,
monotonicity on rays, null at infinity, etc. We also focus on the projection
depth in this study because of these desirable properties.  Furthermore,
authors in the previous work studied an extension of the spatial depth-based
outlier detection with kernel method~\cite{Chen2009}. The other study proposed
an extension of Mahalanobis depth in the reproducing kernel Hilbert
space~\cite{Yonggang2011}. Their studies are also related to ours because we
consider an extension of the projection depth with the kernel method for
outlier detection. In the following, we will describe our motivation in detail.

We have examined the performance of outlier detection methods based on Random
Projection Outlyingness (RPO). The detection performance will degrade when data
samples are not linearly separable, and the sample distribution exhibits
multiple modalities. To cope with the such modalities, we have proposed the
RPO-based detection method with a local score
weighting~\cite{tamamori}. However, the detection performance still needs to
improve since it cannot cope with a nonlinearly separable data cloud. The RPO
involves inner product computations, i.e., linear operations in the original
feature space. In contrast, kernel-based detection methods such as One-Class
Support Vector Machine~(OCSVM)~\cite{OCSVM2001} can deal with such nonlinearly
separable distributions using a feature map into Reproducing Kernel Hilbert
Space (RKHS), and the kernel function induces the inner product in the
RKHS. Combining kernel methods and projection depth will open the possibility
of coping with nonlinear separability.

Given a large amount of training data available, we can use detection methods
based on deep learning, such as Deep Support Vector Data Description~(Deep
SVDD)~\cite{DeepSVDD2018} and Deep AutoEncoding Gaussian Mixture
Model~\cite{DAGMM}. These methods first perform a nonlinear mapping which maps
the original features into a new feature space, and the detector is trained on
the new space. Especially concerning RPO, Deep RPO~\cite{DeepRPO2021} has been
proposed; it combines Deep SVDD and RPO in the training objective. The authors
demonstrated that it could improve detection performance than Deep SVDD and
RPO. However, training such models on a small dataset will be challenging
because sufficient data samples are only sometimes available.

This study proposes an outlier detection method based on Kernel Random
Projection Depth (KRPD). A reproducing kernel function maps original feature
vectors into the RKHS, and we can compute RPD on the RKHS. Although authors
have proposed several methods to obtain a nonlinear mapping related to kernel
approximation~\cite{random_fourier, Nystroem}, we adopt Kernel Principal
Component Analysis~(KPCA) to reduce the dimensionality of the feature
vector. This dimensionality reduction helps make the algorithm feasible and
avoids the curse of dimensionality related to the computations of an inner
product in the RKHS. We conducted experiments on benchmark datasets for outlier
detection. The results demonstrated that the proposed method outperformed RPD
and performed better than other competitive detection methods, including OCSVM,
Local Outlier Factor~\cite{Breunig2000}, Isolation Forest~\cite{IForest2008},
and Gaussian Mixture Model~\cite{Barkan2016}, in terms of Area Under the
Curves~(AUCs) of Receiver Operating Characteristic~(ROC).

\section{Random Projection Depth}
Suppose we are given a $d$-variate probability distribution $P$ and
a data cloud consisting of $N$ data samples
\begin{equation}
  \label{eq:3}
  X = \{\mathbf{x}_1, \mathbf{x}_2,\ldots, \mathbf{x}_N\},\; \;\mathbf{x}_n \sim P,
  \; \;\mathbf{x}_n \in \mathbb{R}^{d}.
\end{equation}
Random Projection Outlyingness (RPO) $O_d(\mathbf{x}\mid X)$ on $X$ is defined as
\begin{equation}
  \label{eq:RPD_d}
  O_d(\mathbf{x}\mid X)
  \coloneq \underset{\mathbf{u} \in \mathbb{S}^{d-1}}{\mathrm{sup}}
  \dfrac{\vert \mathbf{u}^T\mathbf{x}
    - \mathrm{MED}(\mathbf{u}^TX) \vert}
  {\mathrm{MAD}(\mathbf{u}^TX)},
\end{equation}
where $\mathbf{x} \in \mathbb{R}^{d}$ is a $d$-dimensional data point and
$\mathbf{u}$ is a random unit vector drawn from the unit hypersphere
$\mathbb{S}^{d-1}$.  The one-dimensional, scalar-valued dataset
$\mathbf{u}^{T}X$ is given by
$\mathbf{u}^TX = \{\mathbf{u}^{T}\mathbf{x}_1, \mathbf{u}^{T}\mathbf{x}_2,
\ldots, \mathbf{u}^{T}\mathbf{x}_N\}$. The abbreviations MED and MAD denote the
median and the median absolute deviation, respectively.  In accordance with the
above definition, we can define Random Projection Depth (RPD)
$D_d(\mathbf{x}\mid X)$~\cite{projection_depth} as
\begin{equation}
  D_d(\mathbf{x}\mid X) \coloneq \frac{1}{1 + O_d(\mathbf{x}\mid X)}.
\end{equation}
Generally, the depth function is a statistic that gives a center-outward
ordering to each data point $\mathbf{x}$.  The point at which the depth
function takes its maximum value is the median, a \textit{deepest} point. RPD
has the characteristic of not being able to deal with multimodal
distributions. When we utilize RPD for outlier detection, this characteristic
will likely decrease detection accuracy and robustness.  To address this
problem, we have proposed a method of weighting the outlyingness based on the
cardinality of the clusters to which the samples belong~\cite{tamamori}.
However, this method requires the number of clusters to be determined in
advance. Another characteristic of RPD is that the central region forms a
convex hull, which also causes an increase in false positive errors when the
data cloud is non-convex.

In order to demonstrate the above issues of RPD, we show contour plots and
decision surfaces across synthetic datasets in Figure~\ref{fig:rpo_datasets}.
In this figure, datasets are: (a) Unimodal, (b) Multimodal, (c) Cross, and (d)
Moons.  We generated each dataset with 400 samples (300 inliers, 100 outliers)
in two-dimensional space. The true outlier samples were generated by a uniform
distribution within $[-6, 6] \times [-6, 6]$.  For Unimodal, the inlier samples
were generated by a scaled standard normal distributions where the scale was
set to $0.3$.  For Multimodal, the inlier samples were generated by three
scaled standard normal distributions.  The scale of each normal distribution
was also the same value as $0.3$.  We generated inliers for Cross on the two
vertical segments and Moons on the shifted semicircles.  Small noises were
added to those inliers; the strength of noise was set to $0.05$ uniformly.

In this study, we utilize a negative projection depth as the outlier score; a
high-depth value corresponds to an inlier, and a low-depth value corresponds to
an outlier.  We can identify a data point as an outlier if its outlier score is
greater than a threshold~\cite{Chen2009}. The threshold was determined as
$25\; (= 100 \times 100 / 400)$ percentile within whole predicted scores.  We
specify the central region as a set of points where its outlier score exceeds
the threshold. The region is colored orange in the figure.  We refer to the
boundary of the central region as the decision boundary.  The central region
should include many inliers and few outliers in practice.  On Unimodal, we can
see that RPD successfully produces a tight boundary and the decision surface
encloses almost all inliers. In this case, we can expect that outlier detection
works well.  However, on other datasets including Multimodal we can see that
RPD fails to form tight decision boundaries to inliers and encloses inliers and
many outliers.  The false positive errors will increase in such cases.

\section{Proposed method}\label{sec:proposed}
We propose to compute RPD in an RKHS to resolve the issues described in the
previous section, where RPD-based detection will fail on the multi-modal or not
lineally separable data cloud.  We refer to RPD in the RKHS as Kernel
RPD~(KRPD).  The kernel method~\cite{Schlkopf} makes the computation of KRPD
feasible.  In the following, we formulate the proposed method.

Suppose we are given a positive definite real-valued kernel
\begin{equation}
  k: \Omega \times \Omega \rightarrow \mathbb{R},
\end{equation}
where $\Omega$ is a non-empty subset of $\mathbb{R}^{d}$.
We refer the RKHS induced by~$k$ as $\mathcal{H}_k$ and define a feature map
\begin{equation}
  \Phi: \Omega \rightarrow \mathcal{H}_k, \; \mathbf{x} \in \Omega
  \mapsto k(\cdot, \mathbf{x}).
\end{equation}
The data cloud $X$ is mapped to the data cloud $\mathcal{F}_{k}(X)$ on $\mathcal{H}_k$:
\begin{equation}
  \mathcal{F}_{k}(X) = \{\Phi(\mathbf{x}_1), \Phi(\mathbf{x}_2), \ldots,
  \Phi(\mathbf{x}_N)\}  \subset \mathcal{H}_k.
\end{equation}
We can define the outlyingness
$\mathcal{O}_{k}(\mathbf{x}\mid \mathcal{F}_{k}(X))$
and the projection depth
$\mathcal{D}_{k}(\mathbf{x}\mid \mathcal{F}_{k}(X))$
on $\mathcal{H}_k$ sequentially as
\begin{equation}
  \label{eq:KRPD}
  \mathcal{O}_{k}(\mathbf{x}\mid \mathcal{F}_{k}(X))
  \coloneq \underset{\|f\| =1}{\mathrm{sup}}
      \dfrac{\vert \langle f, \Phi(\mathbf{x}) \rangle
      - \mathrm{MED}( \langle f,\mathcal{F}_{k}(X) \rangle ) \vert}
    {\mathrm{MAD}(\langle f,\mathcal{F}_{k}(X) \rangle)},
\end{equation}
and
\begin{equation}
  \mathcal{D}_{k}(\mathbf{x} \mid \mathcal{F}_{k}(X))
  \coloneq \frac{1}{1 + \mathcal{O}_k(\mathbf{x}\mid~\mathcal{F}_{k}(X))},
\end{equation}
where $\|\cdot \|$ denotes the norm in $\mathcal{H}_k$.
The $\langle f,\mathcal{F}_{k}(X) \rangle$ is an abbreviation of
the scalar-valued data cloud
\begin{equation}
 \langle f,\mathcal{F}_{k}(X) \rangle
 = \{
  \langle f, \Phi(\mathbf{x}_1) \rangle,
  \langle f, \Phi(\mathbf{x}_2) \rangle,
  \ldots,
  \langle f, \Phi(\mathbf{x}_N) \rangle\},
\end{equation}
where $\langle \cdot, \cdot \rangle$ denotes the inner product of
$\mathcal{H}_k$. Through nonlinear kernel function $k$ and its feature map, we
expect that the projection depth is robust to the distribution with multi
modality and non-convexity.

Since the dimensionality of $\mathcal{H}_k$ is not necessarily finite, the unit
sphere $\mathcal{S}_k = \{f \in \mathcal{H}_k \mid \|f\| =1 \}$ is generally
not a compact set. This leads to the case that the supremum
of~Eq.(\ref{eq:KRPD}) cannot be approximated through a sequence of elements in
$\mathcal{S}_k$ and the computation of the projection depth becomes infeasible.
To derive the computationally feasible algorithm, we resort to the representer
theorem~\cite{Schlkopf}. We can assume that an element $f \in \mathcal{S}_k$
belongs to the linear span of $\mathcal{F}_{k}(X)$:
\begin{equation}
  f  = \sum_{n=1}^{N} \alpha_{n} \Phi(\mathbf{x}_n)
 = \sum_{n=1}^{N} \alpha_{n} k(\cdot, \mathbf{x}_n),
 \label{eq:representer}
\end{equation}
where $(\alpha_1, \alpha_2, \ldots, \alpha_N) \in \mathbb{R}^{N}.$
The condition $\|f\| =1$ can be rewritten as
\begin{equation}
  \label{eq:unit_sphere}
  \boldsymbol{\alpha}^{T} \mathbf{K} \boldsymbol{\alpha} = 1,
\end{equation}
where $\boldsymbol{\alpha} = (\alpha_1, \alpha_2, \ldots, \alpha_N)^{T}$
and $\mathbf{K}$ is the $N \times N$ Gram matrix:
\begin{equation}
  \label{eq:kernel_matrix}
  \mathbf{K} =
  \begin{pmatrix}
    k(\mathbf{x}_1, \mathbf{x}_1) & k(\mathbf{x}_1, \mathbf{x}_2)
    & \cdots  & k(\mathbf{x}_1, \mathbf{x}_N)\\
    k(\mathbf{x}_2, \mathbf{x}_1) & k(\mathbf{x}_2, \mathbf{x}_2)
    & \cdots  & k(\mathbf{x}_2, \mathbf{x}_N)\\
    \vdots & \vdots & \ddots & \vdots\\
    k(\mathbf{x}_N, \mathbf{x}_1) & k(\mathbf{x}_N, \mathbf{x}_2)
    &\cdots  & k(\mathbf{x}_N, \mathbf{x}_N)
  \end{pmatrix}.
\end{equation}

We can indeed compute the projection depth approximately in $\mathcal{H}_k$ by
sampling $\boldsymbol{\alpha}$ from the hyperellipsoid
of~$\boldsymbol{\alpha}^{T} \mathbf{K} \boldsymbol{\alpha} = 1$. However, the
inner product between high dimensional vectors suffers from the curse of
dimensionality; almost all vectors in a high dimensional space are orthogonal
to each other~\cite{Hall2005, wainwright2019}.  This characteristic will
significantly reduce the efficiency and accuracy of depth computation.  To
improve the efficiency and accuracy, we propose to perform a dimensionality
reduction by KPCA~\cite{Scholkopf1998} before computing projection depth.  In
this paper, we denote the principal dominant components as $\boldsymbol{\beta}$
for a data point $\mathbf{x}$ and its dimensionality as $M$.  We also denote
the new data cloud as
$F(X) = \{\boldsymbol{\beta}_1, \boldsymbol{\beta}_2, \ldots,
\boldsymbol{\beta}_N\}$.  By treating $\boldsymbol{\beta}$ and
$\boldsymbol{\beta}_n$ as new coordinates in $\mathbb{R}^{M}$, the projection
depth can be computed in the same manner as in $\mathbb{R}^{M}$.  If setting
$M < N$, we can simultaneously achieve a nonlinear dimensionality reduction and
a feasible computation of the projection depth.  The overall computation
algorithm is summarized in Algorithm~\ref{alg:cap}.
\begin{algorithm}[t]
  \caption{Computation of outlier score based on KRPD.}\label{alg:cap}
  \begin{algorithmic}[1]
    \Require Data cloud $X = \{\mathbf{x}_1, \mathbf{x}_2, \ldots, \mathbf{x}_N\}
    \subset \Omega$, kernel function~$k$, dominant
    dimensionality $M$, the number of random projection vectors $L$,
    and a data point $\mathbf{x} \in \Omega$.
    \Ensure Negative projection depth $-D_{M}(\boldsymbol{\beta} \mid F(X))$.
    \State Compute the Gram matrix $\mathbf{K}$ between $\mathbf{x}$ and $X$,
    and obtain the centralized matrix $\mathbf{K}^{\prime}$ from $\mathbf{K}$~\cite{Schlkopf}.
    \State Compute principal
    components $\boldsymbol{\beta}_n \in \mathbb{R}^{M}$ for each $\mathbf{x}_n$
    by solving the eigenvalue problem of $\mathbf{K}^{\prime}$,
    and obtain the new data cloud: $F(X) = \{\boldsymbol{\beta}_1,
    \boldsymbol{\beta}_2, \ldots, \boldsymbol{\beta}_N\}$.
    \State Draw a set of $L$ random unit vectors from the unit hypersphere $\mathbb{S}^{M-1}$:
    $U = \{\mathbf{u}_1, \mathbf{u}_2 \ldots, \mathbf{u}_L \} \subset \mathbb{S}^{M-1}$.
    \State Compute the principal component $\boldsymbol{\beta} \in \mathbb{R}^{M}$
    for $\mathbf{x}$.
    \State Compute the projected data cloud $\mathbf{u}_l^TF(X)$
    for $1 \leq l \leq L$.
    \State \Return $-D_{M}(\boldsymbol{\beta} \mid F(X))$ in accordance with
    Eq.~(\ref{eq:KRPD_approx}).
  \end{algorithmic}
\end{algorithm}
In the step~5), the projected data cloud $\mathbf{u}_l^TF(X)$ is defined as
$\mathbf{u}_l^TF(X) \coloneq
\{\mathbf{u}_l^T\boldsymbol{\beta}_1,
\mathbf{u}_l^T\boldsymbol{\beta}_2, \ldots, \mathbf{u}_l^T\boldsymbol{\beta}_N^{T}\}$.
The projection depth~$\mathcal{D}_{k}(\mathbf{x}\mid~\mathcal{F}_{k}(X))$ is approximated as
\begin{align}
  \hspace{-2mm}\mathcal{D}_{k}(\mathbf{x}\mid \mathcal{F}_{k}(X))
  &\approx D_{M}(\boldsymbol{\beta} \mid F(X))\\
  &\approx \left(
      1 + \underset{1 \leq l \leq L}{\mathrm{max}}
      \dfrac{\vert \mathbf{u}_l^{T}\boldsymbol{\beta}
      - \mathrm{MED}(\mathbf{u}_l^TF(X)) \vert}
      {\mathrm{MAD}(\mathbf{u}_l^TF(X))}
      \right)^{-1}. \label{eq:KRPD_approx}
\end{align}
We also utilize a negative projection depth as the outlier score in the same
way as RPD. The outlier detection algorithm is also the same.

\begin{figure*}[t]
  \centering
  \begin{subfigure}{\linewidth}
    \centering
    \includegraphics[width=\linewidth]{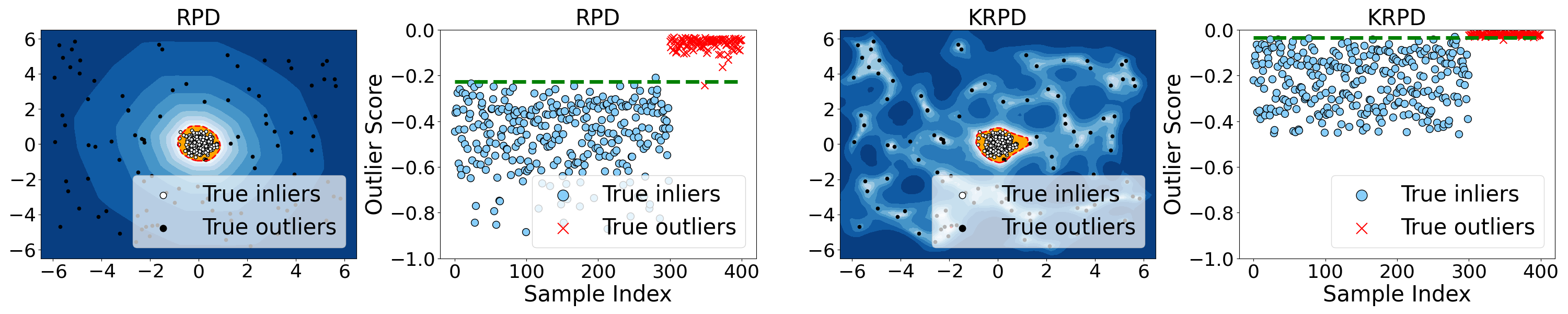}
    \caption{Unimodal}
    \label{fig:score_unimodal}
  \end{subfigure}
  \\
  \begin{subfigure}{\linewidth}
    \centering
    \includegraphics[width=\linewidth]{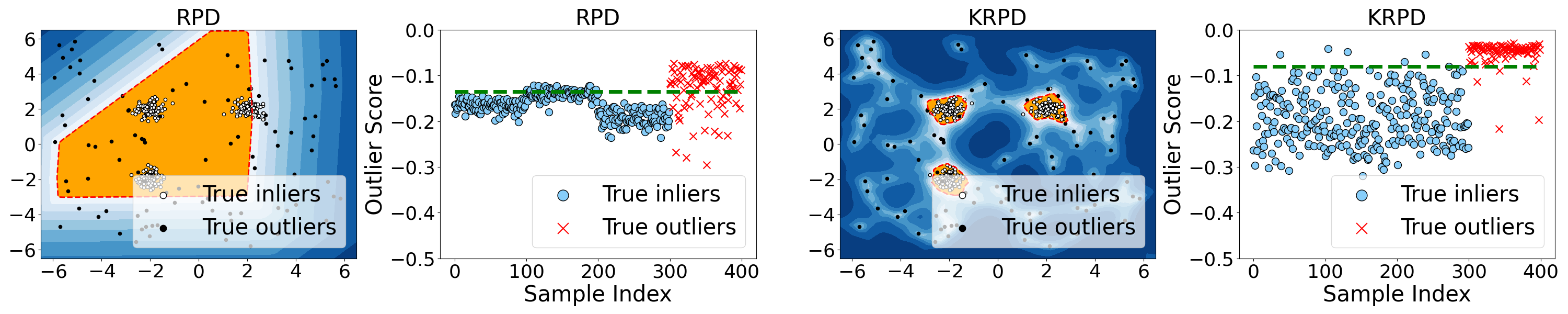}
    \caption{Multimodal}
    \label{fig:score_multimodal}
  \end{subfigure}
  \\
  \begin{subfigure}{\linewidth}
    \centering
    \includegraphics[width=\linewidth]{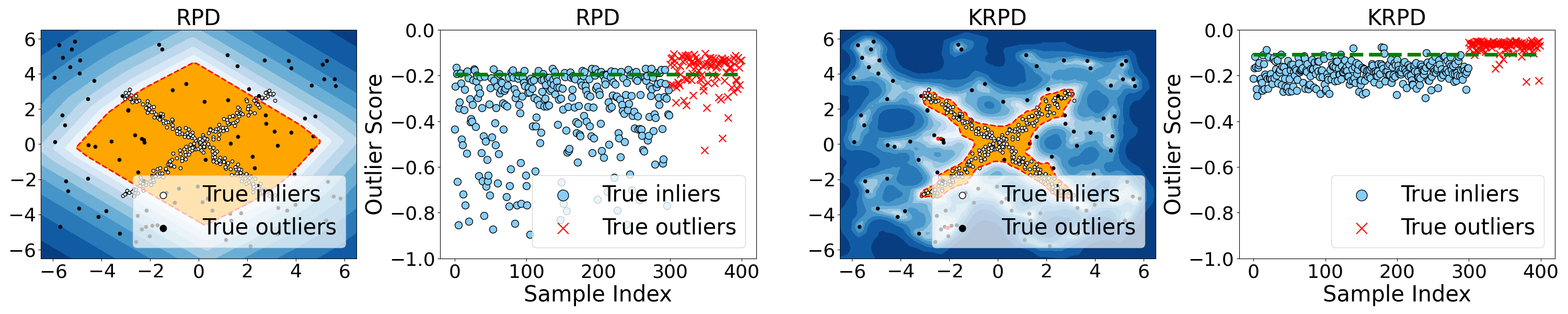}
    \caption{Cross}
    \label{fig:score_cross}
  \end{subfigure}
  \\
  \begin{subfigure}{\linewidth}
    \includegraphics[width=\linewidth]{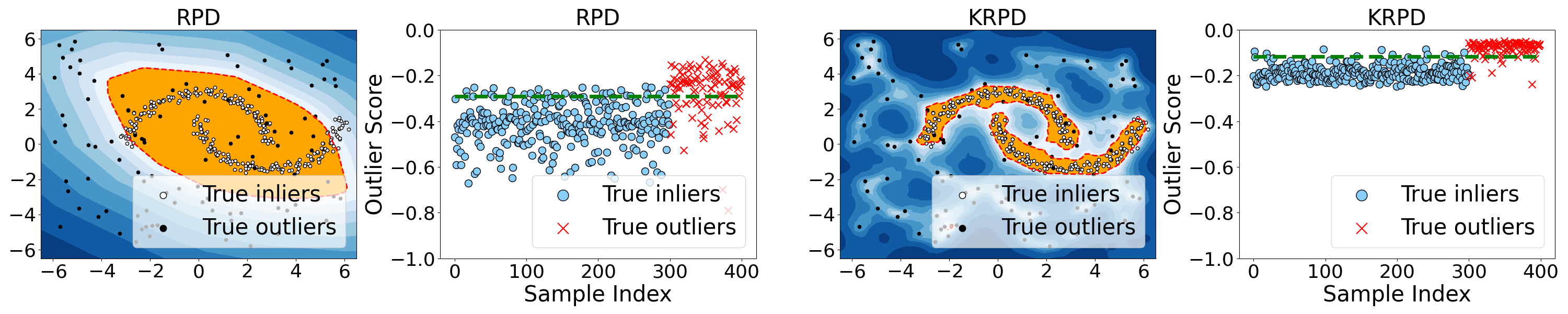}
    \caption{Moons}
    \label{fig:score_moons}
  \end{subfigure}
  \caption{Contour plots and outlier score comparison between RPD and KRPD.
    The threshold is indicated by the green dotted line in the outlier score
    plot. High outlier score corresponds to strong outlyingness.  }
  \label{fig:score_comparison}
\end{figure*}

\section{Experimental Evaluation}
We conducted experiments on toy example data and benchmark datasets to evaluate
the detection performance of proposed model.  The following detection methods
were compared with KRPD:
\begin{itemize}
\item $\mathit{k}$-Nearest Neighbors ($\mathit{k}$NN)~\cite{Angiulli2002}
\item Local Outlier Factor (LOF)~\cite{Breunig2000}
\item Isolation Forest (IForest)~\cite{IForest2008}
\item One-Class Support Vector Machine (OCSVM)~\cite{OCSVM2001}
\item Kernel Principal Component Analysis (KPCA)~\cite{KernelPCA}
\item Random Projection Depth (RPD)~\cite{projection_depth}
\end{itemize}
In this study, RBF kernel function
$k(\mathbf{x}, \mathbf{y}) = \exp(-\gamma || \mathbf{x} - \mathbf{y}||^{2})$
was used for OCSVM, KPCA and KRPD.

\subsection{Toy Example}
\begin{figure*}[t]
  \centering
  \includegraphics[width=\linewidth]{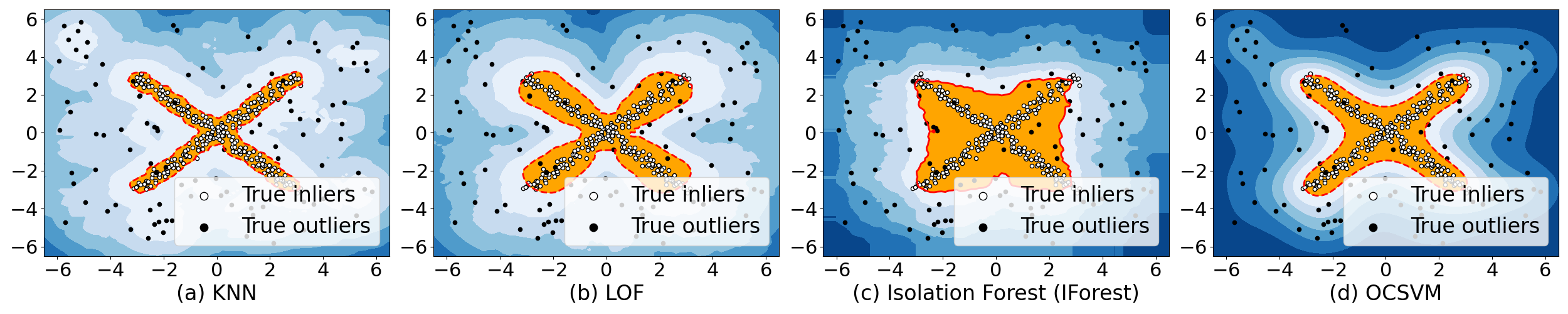}
  \caption{Contour plots of detection methods on the Cross dataset.}
  \label{fig:compare_models_cross}
\end{figure*}
\begin{figure*}[t]
  \centering
  \includegraphics[width=\linewidth]{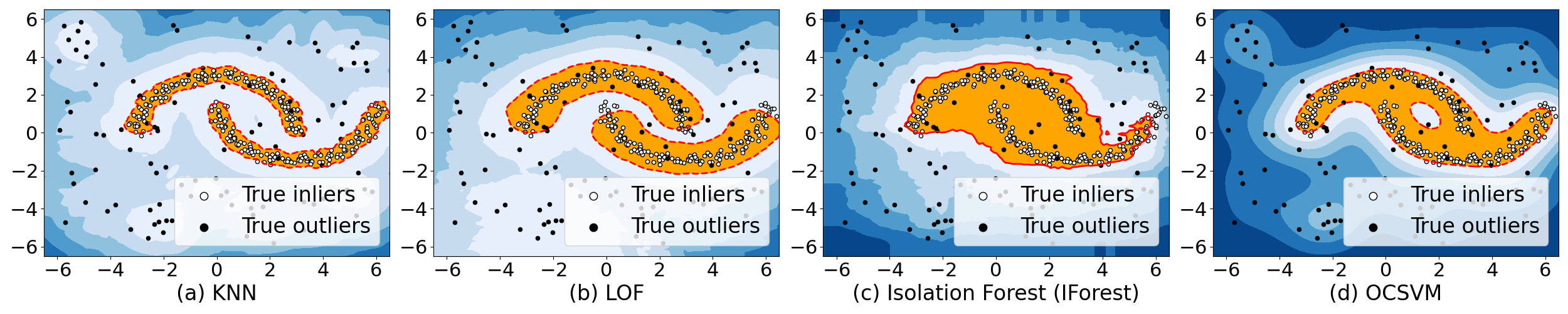}
  \caption{Contour plots of detection methods on the Moon dataset.}
  \label{fig:compare_models_moons}
\end{figure*}
\begin{figure*}[t]
  \centering
  \includegraphics[width=\linewidth]{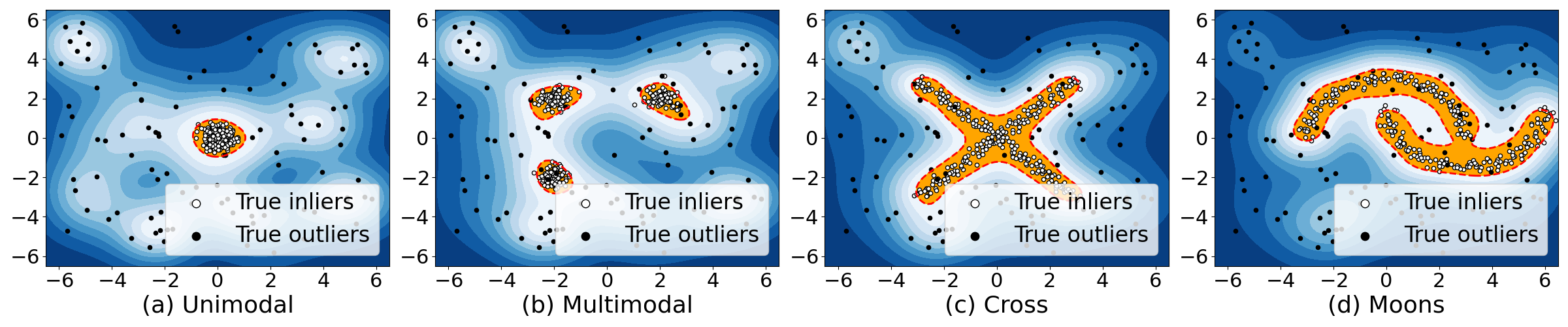}
  \caption{Contour plots of KPCA-based detector on the toy datasets.}
  \label{fig:kpca_datasets}
\end{figure*}

Figure~\ref{fig:score_comparison} shows contour plot comparisons between RPD
and KRPD across toy datasets.  We used the same datasets as in
Figure~\ref{fig:rpo_datasets}.  We fixed the kernel parameter $\gamma$ and the
dominant dimensionality $M$ to $0.25$ and $100$, respectively.  The threshold
of the central region was determined according to the ratio of inliers to the
total samples. Namely, the 75 ($= 100 \times 300 / 400$) percentile of the
outlier scores. We also show the outlier scores on each toy data dataset. In
these figures, the central region is colored orange and the threshold is
represented as the green dotted line. First, we can see that the decision
boundaries of both RPD and KRPD could enclose inliers on the unimodal. It is
apparent that the corresponding scores are separated well, and consequently, we
obtain few detection errors. Compared with RPD, we can see that the decision
boundary of KRPD successfully encloses almost all inlier samples, and many
outliers are outside the central region even when the data cloud exhibits
multiple modalities and non-convexity. On the multimodal, scores of RPD from
several inliers exceed the threshold. Such exceeded scores lead to an increase
in false negative detection errors.  Furthermore, many scores of RPD from
outliers are lower than the threshold, which leads to an increase in false
positive errors. In contrast, both false positive/negative detection errors
decrease in KRPD, as we can see from the right side of the figure. The trend is
also generally true for the remaining Cross and Moons. From these results, we
can expect that KRPD will demonstrate better detection performance than RPD.

Figure~\ref{fig:compare_models_cross} and \ref{fig:compare_models_moons} show
the contour plot comparisons on the Cross and Moons dataset, respectively. We
can see that the decision boundary of IForest circumscribed the dataset and
failed to capture the modalities, which led to an increase in false positive
errors. $\mathit{k}$NN, LOF, and OCSVM achieved tighter boundaries to capture
the modalities.  Figure~\ref{fig:kpca_datasets} shows the contour plots of KPCA
on the toy datasets. We applied the same RBF kernel and the hyperparameter
$\gamma$ for both KPCA and KRPD. The dominant dimensionality $M$ was set to 10
for KPCA.  In the KPCA-based detection, we used the reconstruction error in the
feature space as the outlier score~\cite{KernelPCA}. From the figure, we can
see that the boundary fitted the data cloud tightly, and even only KPCA
obtained favorable feature spaces for outlier detection.  Comparing with these
figures and Figure~\ref{fig:score_comparison}, we consider that KRPD could
achieve the competitive decision boundaries to other methods.

\subsection{Performance Evaluation on Benchmark Datasets}
\begin{table}[t]
  \caption{
    Information of benchmark dataset~\cite{Rayana2016}: dataset name,
    number of samples, number of feature dimensions, and outlier percentages.}
  \label{tab:ODDS}
  \centering
  \begin{tabular}{c c c c}
    \toprule
    Dataset & \#samples & \#dims. & outliers (\%)\\
    \midrule
    Arrhythmia & 452 & 274 & 14.6\% \\
    Cardio & 1831 & 21 & 9.61\% \\
    Ionosphere & 351 & 33 & 35.9\% \\
    Letter & 1600 & 32 & 6.25\% \\
    MNIST & 7603 & 100 & 9.21\% \\
    Musk & 3062 & 166 & 3.17\% \\
    Optdigits & 5216 & 64 & 2.88\%\\
    Pendigits & 6870 & 16 & 2.27\%\\
    Satellite & 6435 & 36 & 31.6\% \\
    Satimage-2 & 5803 & 36 & 1.22\%\\
    Vowels & 1456 & 12 & 3.43\% \\
    Wbc & 378 & 30 & 5.56\%\\
    \bottomrule
  \end{tabular}
\end{table}

\begin{table*}[t]  \centering
  \caption{ROC AUC scores of outlier detection methods.
    Average of 5 independent trials is shown. The better method is indicated in
    \textbf{bold}.} 
  \label{table:approx_roc}
  \begin{tabular}
    {c |c c c c c c c}
    \toprule 
    \textbf{Dataset}
    & \textbf{$\mathit{k}$NN}
    & \textbf{LOF}
    & \textbf{IForest}
    & \textbf{OCSVM}
    & \textbf{KPCA}
    & \textbf{RPD}
    & \textbf{KRPD (ours)} \\
    \cmidrule{2-8}
    Arrhythmia
    & 0.783 $\pm$ 0.024
    & 0.774 $\pm$ 0.028
    & \textbf{0.796 $\pm$ 0.040}
    & 0.781 $\pm$ 0.022
    & 0.768 $\pm$ 0.026
    & 0.771 $\pm$ 0.021
    & 0.791 $\pm$ 0.026 \\
    Cardio
    & 0.841 $\pm$ 0.018
    & 0.741 $\pm$ 0.017
    & 0.928 $\pm$ 0.014
    & 0.924 $\pm$ 0.012
    & 0.852 $\pm$ 0.029
    & \textbf{0.936 $\pm$ 0.017}
    & 0.926 $\pm$ 0.015 \\
    Ionosphere
    & 0.924 $\pm$ 0.016
    & 0.889 $\pm$ 0.030
    & 0.839 $\pm$ 0.021
    & 0.910 $\pm$ 0.017
    & 0.918 $\pm$ 0.019
    & 0.866 $\pm$ 0.014
    & \textbf{0.935 $\pm$ 0.012} \\
    Letter
    & \textbf{0.897 $\pm$ 0.011}
    & 0.865 $\pm$ 0.016
    & 0.614 $\pm$ 0.029
    & 0.792 $\pm$ 0.028
    & \textbf{0.897 $\pm$ 0.018}
    & 0.617 $\pm$ 0.022
    & 0.865 $\pm$ 0.018
    \\
    MNIST
    & 0.872 $\pm$ 0.004
    & 0.779 $\pm$ 0.008
    & 0.785 $\pm$ 0.008
    & 0.859 $\pm$ 0.037
    & \textbf{0.875 $\pm$ 0.004}
    & 0.844 $\pm$ 0.013
    & 0.856 $\pm$ 0.006 \\
    Musk
    & 0.972 $\pm$ 0.009
    & 0.569 $\pm$ 0.028
    & \textbf{0.998 $\pm$ 0.000}
    & 0.836 $\pm$ 0.085
    & 0.991 $\pm$ 0.011
    & \textbf{0.998 $\pm$ 0.000}
    & 0.997 $\pm$ 0.005 \\
    Optdigits
    & 0.394 $\pm$ 0.028
    & 0.605 $\pm$ 0.080
    & 0.686 $\pm$ 0.043
    & 0.528 $\pm$ 0.041
    & 0.585 $\pm$ 0.053
    & 0.552 $\pm$ 0.042
    & \textbf{0.755 $\pm$ 0.070}\\
    Pendigits
    & 0.842 $\pm$ 0.005
    & 0.521 $\pm$ 0.025
    & 0.950 $\pm$ 0.013
    & 0.870 $\pm$ 0.038
    & \textbf{0.953 $\pm$ 0.010}
    & 0.838 $\pm$ 0.029
    & 0.920 $\pm$ 0.008 \\
    Satellite
    & 0.717 $\pm$ 0.009
    & 0.569 $\pm$ 0.007
    & 0.714 $\pm$ 0.018
    & 0.715 $\pm$ 0.039
    & 0.716 $\pm$ 0.007
    & 0.667 $\pm$ 0.004
    & \textbf{0.727 $\pm$ 0.005} \\
    Satimage-2
    & 0.992 $\pm$ 0.001
    & 0.700 $\pm$ 0.057
    & 0.989 $\pm$ 0.004
    & 0.963 $\pm$ 0.033
    & 0.993 $\pm$ 0.004
    & \textbf{0.996 $\pm$ 0.002}
    & 0.994 $\pm$ 0.006 \\
    Vowels
    & \textbf{0.966 $\pm$ 0.009}
    & 0.934 $\pm$ 0.009
    & 0.757 $\pm$ 0.040
    & 0.910 $\pm$ 0.049
    & 0.940 $\pm$ 0.040
    & 0.810 $\pm$ 0.049
    & 0.905 $\pm$ 0.021 \\
    Wbc
    & 0.812 $\pm$ 0.026
    & 0.915 $\pm$ 0.021
    & \textbf{0.937 $\pm$ 0.028}
    & 0.881 $\pm$ 0.062
    & 0.884 $\pm$ 0.038
    & 0.911 $\pm$ 0.032
    & 0.906 $\pm$ 0.012 \\
    \midrule
    \textbf{Average}
    & 0.834 $\pm$ 0.013
    & 0.738 $\pm$ 0.027
    & 0.832 $\pm$ 0.022
    & 0.831 $\pm$ 0.039
    & 0.864 $\pm$ 0.022
    & 0.817 $\pm$ 0.020
    & \textbf{0.882 $\pm$ 0.017}\\
    \midrule
  \end{tabular}
\end{table*}

We conducted experiments on twelve standard benchmark datasets publicly
available at ODDS library~\cite{Rayana2016} for performance evaluations. The
detailed information is listed in Table~\ref{tab:ODDS}. We randomly split each
dataset into training and test data in a stratified fashion, specifying a seed;
60\% for training and 40\% for testing. The outlier percentage was kept the
same on both training and test because of the stratified fashion. This split
scheme intends to simulate data contamination and evaluate the robustness of
the existence of outliers. Moreover, we performed a 5-fold stratified
cross-validation on the training data for hyperparameter search. In each fold,
80\% of training data were used for actual model training, and the rest of 20\%
were used for validation.

For both RPD and KRPD, the number of axes $L$ for random projections was fixed
to 1000.  In order to optimize other hyperparameters, we utilized a framework
using the Optuna package~\cite{optuna_2019}, which implements a Bayesian
optimization. In this framework, we selected a strategy to maximize the ROC
AUCs.  For OCSVM, KPCA and KRPD, the kernel parameter $\gamma$ was optimized
between $[0.00001, 1.0]$. For both KPCA and KRPD, the number of principal
components $M$ was optimized between $[10, 500]$.
We can obtain the label information as binary ones because the
training data contains inliers and outliers. Hence we could also compute ROC
AUCs on each fold of the training data, and the hyperparameters were determined
semi-supervised. The Optuna optimized the hyperparameters according to the
maximization criterion.

Table~\ref{table:approx_roc} lists ROC AUCs. We can see that KRPD outperformed
RPD regarding ROC-AUCs on the nine datasets. Significantly, KRPD greatly
improved the detection performance on Letter and Optdigits datasets. On the
three datasets of Cardio, Musk, and Satimage-2, the performance of RPD was
almost the same as KRPD.  The averaged AUCs of KPCA was $0.864\pm 0.022$, and
it was comparable to other detection methods. Although KPCA outperformed KRPD
on several datasets, KRPD outperformed KPCA on average, and especially, the
improvement was significant on the Optdigits.

\section{Discussion}\label{sec:discussion}
\begin{figure*}[t]
  \centering
  \begin{subfigure}{0.49\linewidth}
    \includegraphics[width=\linewidth]{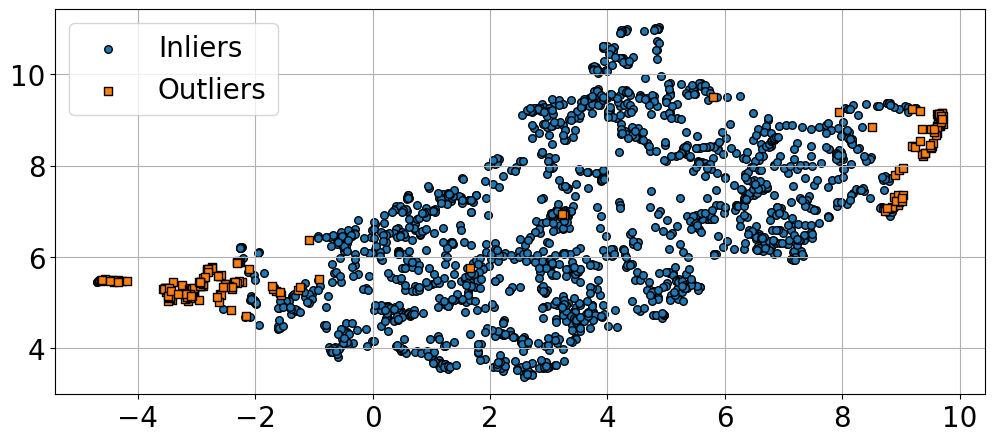}
    \caption{Cardio}
    \label{fig:umap_cardio}
  \end{subfigure}
  \begin{subfigure}{0.49\linewidth}
    \includegraphics[width=\linewidth]{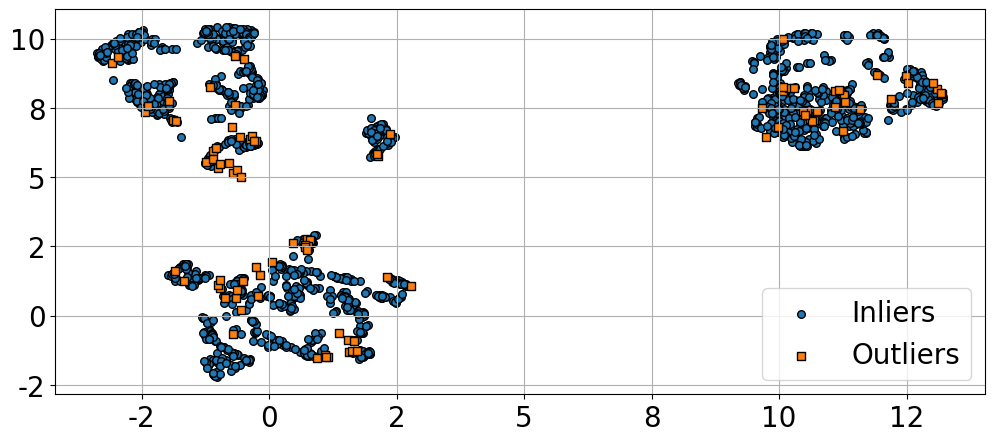}
    \caption{Letter}
    \label{fig:umap_letter}
  \end{subfigure}
  \\
    \begin{subfigure}{0.49\linewidth}
    \includegraphics[width=\linewidth]{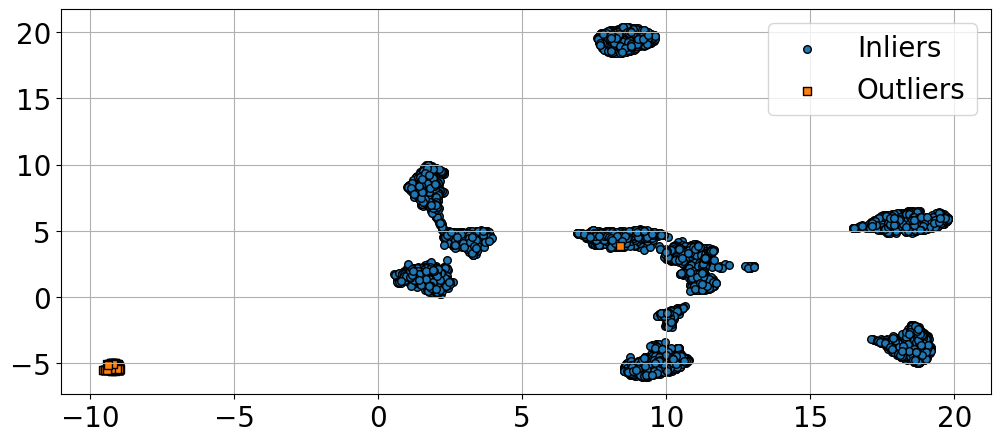}
    \caption{Optdigits}
    \label{fig:umap_optdigits}
  \end{subfigure}
  \begin{subfigure}{0.49\linewidth}
    \includegraphics[width=\linewidth]{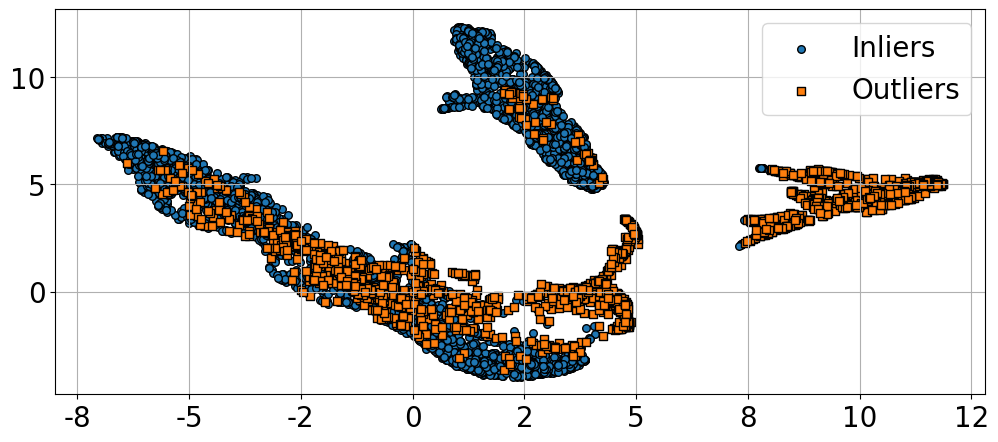}
    \caption{Satellite}
    \label{fig:umap_satellite}
  \end{subfigure}
  \caption{Visualization of datasets by UMAP.}
  \label{fig:umap_visual}
\end{figure*}

We have shown in the previous section that KRPD outperformed RPD and was
comparable to other detection methods on average. These results demonstrate the
effectiveness of KRPD in detecting outliers. However, KRPD did not deliver the
best detection performance across all datasets. In this section, we will
consider why we obtained variations across datasets in the detection
performance of KRPD by using a visualization of datasets.  We will also clarify
the advantages against other methods and limitations of KRPD.

\subsection{Variations of detection performance across datasets}
Figure~\ref{fig:umap_visual} shows visualizations of Cardio, Letter, Optdigits,
and Satellite datasets by UMAP~\cite{umap_arxiv}.  As shown in
Figure~\ref{fig:umap_cardio} of Cardio, almost outliers locate around the edge
of the inlier distribution, and the distribution looks roughly uni-modal. This
observation explains why RPD outperformed KRPD slightly.  Next, on Letter and
Optdigits, KRPD outperformed RPD significantly. From the corresponding
visualization of Letter as shown in Figure~\ref{fig:umap_letter}, we can
observe that roughly three clusters exist, corresponding to three capital
letters in the English alphabet. Those clusters are further split into many
small groups, and outliers are scattered. The detection with KRPD worked well
even in such a challenging dataset. The Optdigits is a collection of instances
of digits 0-9, and digit 0 is the outlier class. We can observe that its data
cloud of Optdigits forms explicit clusters corresponding to each digit, and
hence it shows multiple modalities. These observations also explain the
superior performance of KRPD to RPD and other methods.  As shown in
Figure~\ref{fig:umap_satellite} of Satellite, the data cloud of outliers
overlaps that of inliers, partly explaining the decrease of AUCs for KRPD.
Namely, this decrease explains that KRPD is sensitive to the existence of
outliers, that is, contamination in the training data, leading to an increase
in the false alarm rate.

\subsection{Detection performance without KPCA}
\begin{table}[t]\centering
  \caption{ROC AUC scores of KRPD w/o and w/ KPCA.
    Average of 5 independent trials is shown.
    The better method is indicated in \textbf{bold}.
  }
  \label{table:KRPD_with_without_KPCA}
  \begin{tabular}
    {c |c c}
    \toprule 
    \textbf{Dataset}
    & \textbf{w/o KPCA (RFF)}
    & \textbf{w/ KPCA} \\
    \cmidrule{2-3}
    Arrhythmia
    & 0.768 $\pm$ 0.033
    & \textbf{0.791 $\pm$ 0.026} \\
    Cardio
    & \textbf{0.943 $\pm$ 0.015}
    & 0.926 $\pm$ 0.015 \\
    Ionosphere
    & 0.855 $\pm$ 0.039
    & \textbf{0.935 $\pm$ 0.012} \\
    Letter
    & 0.618 $\pm$ 0.018
    & \textbf{0.865 $\pm$ 0.018} \\
    MNIST
    & 0.820 $\pm$ 0.033
    & \textbf{0.856 $\pm$ 0.006} \\
    Musk
    & \textbf{0.998 $\pm$ 0.002}
    & 0.997 $\pm$ 0.005 \\
    Optdigits
    & 0.543 $\pm$ 0.043
    & \textbf{0.755 $\pm$ 0.070} \\
    Pendigits
    & 0.896 $\pm$ 0.020
    & \textbf{0.920 $\pm$ 0.008} \\
    Satellite
    & 0.689 $\pm$ 0.020
    & \textbf{0.727 $\pm$ 0.005} \\
    Satimage-2
    & \textbf{0.996 $\pm$ 0.003}
    & 0.994 $\pm$ 0.006 \\
    Vowels
    & 0.787 $\pm$ 0.035
    & \textbf{0.905 $\pm$ 0.021} \\
    Wbc
    & 0.896 $\pm$ 0.030
    & \textbf{0.906 $\pm$ 0.012} \\
    \midrule
    \textbf{Average}
    & 0.817 $\pm$ 0.025
    & \textbf{0.882 $\pm$ 0.017} \\
    \midrule
  \end{tabular}
\end{table}

As described in Section~\ref{sec:proposed}, KPCA is applied to avoid reducing
the efficiency and accuracy of depth computation.  Hence, a natural question
arises: How much will the performance degrade without KPCA?  We conducted
additional experiments to investigate the impact of using or not using KPCA on
detection performance.  We selected a kernel approximation method with random
Fourier features~(RFF)~\cite{random_fourier}.  The RFF obtains a high
dimensional vector for each $\mathbf{x}_n$ in the data cloud $X$, and hence we
can compute the projection depth. The number of Monte Carlo samples for RFF,
corresponding to the dimensionality of the feature vector, was optimized
between $[10, 500]$ with the Optuna package. The kernel parameter was also
optimized between $[0.00001, 1.0]$.

Table~\ref{table:KRPD_with_without_KPCA} lists the ROC AUCs of KRPD without and
with KPCA. We can see that the average performance significantly decreased
without KPCA. We consider that this performance degradation was due to the
curse of dimensionality because RFF involves computations of the inner product
in a high dimensional space to approximate the kernel function.  As described
in Section~\ref{sec:proposed}, almost all pairs of vectors on a unit
hypersphere will become orthogonal. This fact significantly affects the
computation of projection depth and partly explains the performance
degradation without KPCA.

\subsection{Limitations and future work}
Since KRPD inherits advantages from KPCA, it also inherits its
limitations. First, careful hyperparameter tuning is often required. In the
experimental evaluation, the KRPD hyperparameters could be determined
semi-supervised since we had labeled samples available. KRPD will be more
effective in semi-supervised situations. Second, KPCA is not necessarily robust
to outliers in the training data. Although KPCA has an application of
denoising~\cite{Mika1998}, its formulation does not assume the existence of
outliers, and the outliers will distort the direction of eigenvectors. Authors
in the previous study~\cite{Wang2020} have proposed several regularization
techniques to obtain eigenvectors robustly against outliers. In the future, we
will consider an extension of KPCA for RPD to cope with outliers and robust
extraction of eigenvectors. Third, the computational complexity is still
$O(N^{3})$, where $N$ is the number of training samples. Approximate kernel
PCA~\cite{Sriperumbudur2022} will be effective when $N$ itself becomes a large
number, e.g., $N > 10^{4}$.  Kernel PCA with an online learning
algorithm~\cite{Honeine2012} will also reduce computational complexity.  We
will investigate correlations among the number of training samples, the
dimensionality of the principal subspace, and its detection performance.
Lastly, we have not given statistical support to explain the improved detection
performance from the viewpoint of statistical depth. We are interested in how
the desirable properties of statistical depth~\cite{Zuo2000} behave when
changing the dimensionality $M$ and the number of training samples $N$.

\section{Conclusion}
In this paper, we proposed KRPD, an extension of RPD for outlier detection. The
original data cloud is mapped to RKHS by a reproducing kernel function, and we
can compute RPD in the RKHS. With the help of KPCA, the decision boundary of
KRPD fits the data cloud tightly even when the cloud shows complicated multiple
modalities and non-convexity. We evaluated outlier detection performance for
RPD, KRPD, and other popular methods on 12 benchmark datasets. The experimental
results demonstrated that KRPD outperformed RPD and was comparable to other
methods. We also discussed the detection performance between KPCA and KRPD and
confirmed that KRPD was more effective than KPCA. We plan to extend KRPD
further to utilize robust KPCA.

\section*{Acknowledgments}
This work was supported by JSPS KAKENHI Grant Number JP21K17771.

\AtNextBibliography{\footnotesize}
\printbibliography

\end{document}